\pdfoutput=1
\documentclass[11pt]{article}

\usepackage{acl}

\usepackage{times}
\usepackage{latexsym}

\usepackage[T1]{fontenc}
\usepackage[utf8]{inputenc}

\usepackage{microtype}

\usepackage{hyperref}

\makeatletter
\newcommand{\printfnsymbol}[1]{%
  \textsuperscript{\@fnsymbol{#1}}%
}
\makeatother

\usepackage{lipsum}
\usepackage{amsmath,amsthm,amssymb}
\usepackage{xspace}
\usepackage[ruled, vlined, algo2e]{algorithm2e}
\usepackage{dsfont}
\usepackage{threeparttable}
\usepackage{enumitem}
\usepackage{subfigure}
\usepackage{graphicx}
\usepackage{booktabs}
\usepackage[skins]{tcolorbox}

\newtcolorbox{myframe}[2][]{%
  enhanced,colback=white,colframe=black,coltitle=black,
  sharp corners,boxrule=0.4pt,
  fonttitle=\itshape,
  attach boxed title to top left={yshift=-0.3\baselineskip-0.4pt,xshift=2mm},
  boxed title style={tile,size=minimal,left=0.5mm,right=0.5mm,
    colback=white,before upper=\strut},
  title=#2,#1
}

\setlist[itemize]{leftmargin=*, topsep=1pt,itemsep=0ex,partopsep=1ex}
\setlist[enumerate]{leftmargin=*, topsep=1pt,itemsep=0ex,partopsep=1ex}

\newcommand{\mpc}{\mathrm{MPC}}

\definecolor{light2}{HTML}{D7F3FD}
\definecolor{light}{HTML}{FFE8D4}
\definecolor{dark}{HTML}{F0997D}
\newcommand{\smallwin}[1]{{\colorbox{light}{#1}}}
\newcommand{\bigwin}[1]{{\colorbox{dark}{{#1}}}}
\newcommand{\threewin}[1]{{\colorbox{light2}{{#1}}}}
\title{Prompted LLMs as Chatbot Modules \\ for Long Open-domain Conversation
}

\author{Gibbeum Lee \textsuperscript{1*}
    \qquad 
    Volker Hartmann \textsuperscript{1*}
    \qquad 
    Jongho Park\textsuperscript{1*} 
    \\ 
    {\bf Dimitris Papailiopoulos \textsuperscript{1,2} 
    \qquad 
    \bf Kangwook Lee \textsuperscript{1,2}} 
    \\
    \vspace{1mm}
    \textsuperscript{\rm 1} KRAFTON~~ 
    \textsuperscript{\rm 2} University of Wisconsin-Madison  \\
    \texttt{\{pirensisco, volker, jongho.park\}@krafton.com}}

\begin{document}
\maketitle
\begingroup\def\thefootnote{*}\footnotetext{Equal Contributions}\endgroup
\begin{abstract}

In this paper, we propose $\mathrm{MPC}$ (Modular Prompted Chatbot), a new approach for creating high-quality conversational agents without the need for fine-tuning. Our method utilizes pre-trained large language models (LLMs) as individual modules for long-term consistency and flexibility, by using techniques such as few-shot prompting, chain-of-thought (CoT), and external memory. Our human evaluation results show that $\mathrm{MPC}$ is on par with fine-tuned chatbot models in open-domain conversations, making it an effective solution for creating consistent and engaging chatbots.
% The increasing reasoning capabilities of large language models (LLMs) have sparked interest in their potential use as chatbots. 
% % However, the process of fine-tuning LMs exclusively on dialogue tasks can be expensive and time-consuming. 
% In this work, we propose $\mathrm{MPC}$ (\textbf{M}odular \textbf{P}rompted \textbf{C}hatbot) which demonstrates the effectiveness of utilizing pretrained LLMs as individual modules for long open-domain conversations. 
% We focus on leveraging these modules to maintain a consistent persona and long-term memory by using techniques such as few-shot prompting, chain-of-thought (CoT), and external memory.
% To this end, we apply various LMs to $\mathrm{MPC}$ and our human evaluation results show that this approach is on par with fine-tuned chatbot models in long-term dialogue.
% Thus, $\mathrm{MPC}$ can be an effective solution for creating a sensible, consistent, and engaging open-domain chatbot.

%  to allow the chatbot to dynamically understand context with the help of external memory. 

\end{abstract}

\section{Introduction}

Language models with billions of parameters, such as GPT-3~\cite{brown2020language} and PaLM~\cite{chowdhery2022palm}, have achieved state-of-the-art performance on many NLP tasks. 
To fine-tune these large language models (LLMs) for open-domain chatbot tasks, one could use a dataset of conversational data that is representative of the target domain. 
However, fine-tuning LLMs for open-domain chatbots can be challenging due to the computational burden of updating models with billions of parameters and the scarcity of data in the dialogue domain.
Furthermore, fine-tuning can limit the model's versatility by restricting it to a specific domain, and result in the loss of domain-agnostic knowledge acquired during pre-training, as reported by~\citet{yang2022improving}.
Multi-task training on different datasets, as proposed by \citet{roller2021recipes}, can address the versatility issue but has limitations, such as the need for data to train each skill and the difficulty determining the necessary skills for an open-domain chatbot. In fact, the growing number of modules for chatbots, as in Blenderbot3 (BB3)~\cite{shuster2022blenderbot}, points towards the increasing burden of data and computation when fine-tuning for each new chatbot model.

Interestingly, some LLMs have the ability to perform in-context learning (ICL)~\cite{nye2022show, wei2022chain, lewkowycz2022solving, wei2022emergent, zhou2022teaching, dasgupta2022language, chung2022scaling}. 
This capability enables the model to rapidly adapt to and execute a specific task based on a brief instruction and a few examples, without requiring additional fine-tuning. 
This can be utilized to create an open-domain chatbot, where a prompt describing a task required for open-domain dialogue and a few examples of solving such task can be provided to the LLM, allowing it to generate information that is pertinent to the current conversation.

\paragraph{Our Contributions}

We present a novel approach for creating high-quality conversational agents without the need for fine-tuning. 
Our proposed chatbot, $\mpc$ (\textbf{M}odular \textbf{P}rompted \textbf{C}hatbot), utilizes open-sourced pre-trained language models to increase the flexibility of designing the modules of an open-domain chatbot.
Our approach enhances multiple conversational capabilities by utilizing a modularized agent that incorporates LLMs with prompt techniques such as few-shot ICL and Chain-of-Thought (CoT).
In the paper, we design MPC to achieve long-term consistency, a domain in which previous chatbots have struggled.
Our human evaluation results show that $\mpc$ is on par with or even preferred over fine-tuned LLMs, such as Blenderbot, in an open-domain conversational setting. 
This approach highlights the potential of pre-trained LLMs to adapt to new tasks without fine-tuning, providing an efficient solution for creating open-domain conversational agents. 
\footnote{Our code is available in \href{https://github.com/krafton-ai/MPC} {https://github.com/krafton-ai/MPC}.}

\section{Related Work}

% \paragraph{LLMs for Generative Tasks}

% ~\citet{kim2022soda} attempts to create large-scale, high-quality open-domain conversation datasets by leveraging LLMs.

\paragraph{Modular Prompting}

Well-crafted elicitive prompts can enhance reasoning abilities, resulting in improved performance across various benchmarks~\cite{kojima2022large, wei2022chain, suzgun2022challenging}.
For complex problems, \citet{press2022measuring} identified the compositionality gap which arises when an LM can solve sub-problems but not the overall solution and further showed that CoT narrows this gap.
Since then, there has been a flurry of work that solves tasks by decomposing them into smaller tasks solved by different ``prompt modules''~\cite{zhou2022teaching, wang2022iteratively, khot2022decomposed, khattab2022demonstrate}.

Modular prompting has found use beyond benchmarks and in conversation generation.
\citet{kim2022soda} used an LLM to generate a socially diverse dialogue dataset that is more natural and detailed than existing crowdsourced datasets. Moreover, hierarchical prompt modules prove to help long-range coherence for generating narratives and plays~\cite{yang2022re3, mirowski2022co}.
We refer to~\citet{mialon2023augmented} for a detailed overview on such augmented uses of LLMs.

\paragraph{Open-domain Chatbots}
Many recent dialogue agents rely on dialogue-finetuned LLMs. In \citet{thoppilan2022lamda}, LaMDa has been trained on large amounts of crawled conversational data and has used a fine-tuned classifier for model safety. 
More recently, similar to our modularization approach, BB3 fine-tunes Open Pre-trained Transformers (OPT) ~\cite{zhang2022opt, shuster2022blenderbot} on QA and dialogue datasets and uses one shared model weight as multiple modules. 

On the other hand, ~\citet{madotto2021few} eliminate the need for fine-tuning on dialogue data by feeding retrieved dialogue samples as few-shot for GPT-J~\cite{gpt-j}. 
We find this work to be complementary to our work, as the few-shot dialogue can be seen as an approach to enhance the utterance generator module.
% However, prompt samples are extracted from the same dataset as is used for evaluation which makes it unclear how well the model will perform in an open-domain setting, especially since human evaluation was not done for FSB.

\paragraph{Long-term Memory}
The Multi-Session Chat dataset~\cite{xu2021beyond} allows for measuring how well conversational agents maintain a long-term memory of facts about the user and bot. Information is retrieved using Dense Passage Retriever (DPR)~\cite{karpukhin2020dense}, while BART compresses memories before storing them. 
In \citet{shuster2022blenderbot}, a modular approach is used to incorporate long-term memory and factual grounding through internet search with an LLM. 
This work is closest to our work since it includes an ablation study in which prompt-based modules are compared with fine-tuned modules. 
However, in our work, we argue that more reasoning-based prompting, as demonstrated in \citet{wei2022chain}, is beneficial for better contextual understanding.

% \paragraph{Evaluation}
% Chatbot evaluations often contain automatic metrics such as perplexity and human evaluation such as SSI~\cite{thoppilan2022lamda}. However, automatic metrics often fail to judge the performance of chatbots since the task can be viewed as a one-to-many problem and especially since the reference response may not be the most appropriate response from a chatbot. Thus, human evaluation still remains to be the best methodology to judge which chatbot is better, though a canonical method remains undetermined~\cite{smith2022human}.
\section{Modular Prompted Chatbot}
\label{sec:modular}

\begin{figure}[h]
\includegraphics[width=0.5\textwidth]{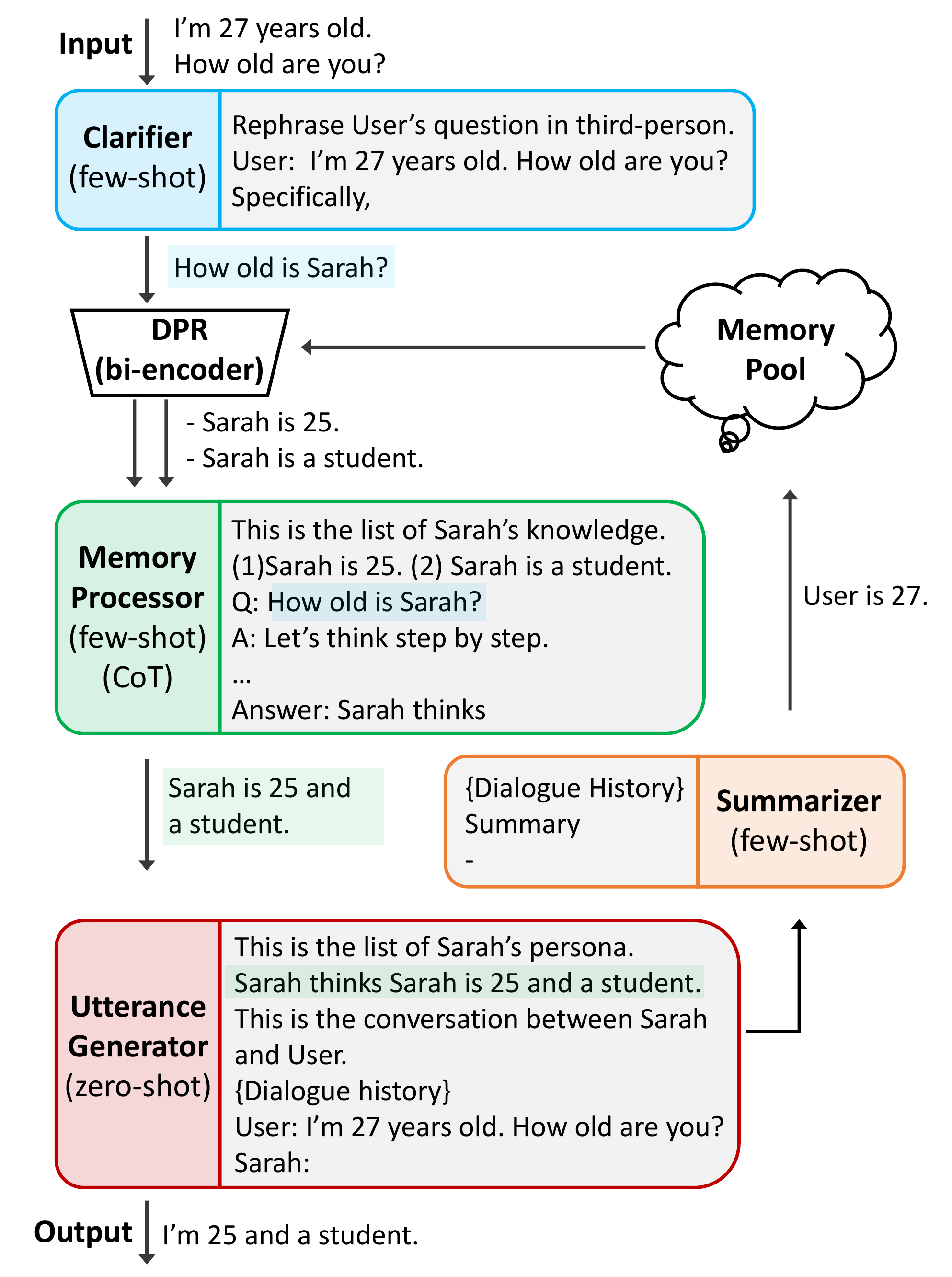}
\caption{Our modular design for improving long-term consistency in open-domain conversation.
}
\label{fig:memory_module}
\centering
\end{figure}

We present a modular chatbot system (Fig.~\ref{fig:memory_module}) that uses prompt-based LLMs to maintain persona and engagement throughout long-term conversations. 

At the start of a conversation, a pre-defined persona is stored in the memory pool.
When a user sends a message, the clarifier rephrases it to resolve any ambiguities and passes it to the DPR model which retrieves relevant memories from the memory pool. 
The retrieved memories and clarifier output are fed into the memory processor to get a single context-relevant memory, which is then passed to an utterance generator for producing a response from the chatbot.
Every few turns, we call upon a summarizer module to extract important information from dialogue and store it in the memory pool for future use (see Appendices~\ref{sec:few_shot_prompts} and~\ref{sec:ug_prompt}). 
% At the start of the conversation, a pre-defined persona (a list of facts) is stored in the memory pool. 
% When a user sends a message to the chatbot, the clarifier rephrases it to resolve any contextual ambiguities. 
% The DPR model then uses this output to retrieve relevant memories from the memory pool. 
% These retrieved memories and clarifier output are fed into the memory processor, which combines them into a single context-relevant memory.  
% This memory is then passed on to an utterance generator for producing an appropriate response from the chatbot. 
% Every few turns, we call upon a summarizer module to extract important information from dialogue and store it in the memory pool for future use.
% We provide the full few-shot prompt for each LM module and examples of utterance generator in Appendix~\ref{sec:few_shot_prompts} and~\ref{sec:ug_prompt}.
% In the following, we explain the purpose and inner workings of each module.

% We initialize the chatbot with a description of its persona, which serves as initial personal history the bot should maintain.

\paragraph{Utterance Clarifier}
As conversations are often muddled with vague coreferences and contextual cues, our clarifier module is an LM prompted with the recent dialogue to resolve any ambiguities. For instance, depending on prior context, the user input ``Do you like working there?'' would output ``Does Sarah like working at ZYX company?''. By resolving contextual ambiguity, the clarifier assists the DPR model and memory processor module by providing an information-dense query to fetch and process relevant memories. 
%The prompt constitutes of few-shot examples with the set of the current utterance of the user and chatbot and the clarified output.

\paragraph{Memory Processor}
% Given the output utterance from the clarifier and multiple relevant memories from the memory pool, the memory processor uses few-shot CoT reasoning to synthesize relevant memories together to produce the most appropriate memory for the context of the conversation. 
As demonstrated in Fig.~\ref{fig:memory_module}, we formulate memory processing as an LLM reasoning task of finding the most relevant information given the dialogue. Following the footsteps in solving hard reasoning tasks~\cite{suzgun2022challenging}, we provide CoT examples to show reasons for ignoring certain memories and synthesizing others. For models incapable of CoT, we simply provide the few-shot examples without the reasoning portion.

Since the memory pool accumulates as the conversation progresses, we use a pre-trained DPR with the output of the clarifier as the query to retrieve the top-$k$ most relevant memories from the memory pool. The memory processor then condenses the top memories into one refined memory.

\paragraph{Utterance Generator}
The utterance generator module generates the final response of the chatbot given the recent dialogue history and memory provided by the memory processor. 
The prompt consists of the dialogue history, condensed memory, and the generation instruction (e.g., ``Give a friendly response to the user.''). 
For some models, we find that inserting the generation instruction at the end was helpful as placing it before the dialogue minimizes the effect of the instruction.

% We find that placing the last few turns after the instruction helps the model to recognize the conversation, while placing it before the entire dialogue minimizes the instruction effect. 
% We provide no dialogue examples in the prompt as few-shot examples often heavily skews generation to undesired output, possibly due to the loss of generalization in terms of text style and dialogue length. 

\paragraph{Dialogue Summarizer}
% The dialogue summarizer module summarizes the conversation history every $T$ turns. 
We provide a few-shot prompt to ensure we record specific details of the conversation and the user. 
% Subsequently, the generated summaries are stored in the memory pool, which consists of the initial description of the persona and the accumulated dialogue summaries.

\section{Experimental Setup}
\label{sec:evaluation}

We evaluate our chatbot's performance by assessing core skills necessary for long-term conversations. 
 We assess consistency by assigning one of five personas, each with 12 facts from PersonaChat ~\cite{zhang2018personalizing}, and presenting these facts to evaluators.
% This setup enables annotators to identify contradictions within a 20-turn conversation window. 
For each experiment, we collect 20 turns from each evaluator and at least 500 turns in total from two subgroups: Amazon Mechanical Turk and university students.
See Appendix~\ref{sec:data_collection} for a detailed explanation of our data collection.

In our setup, there are four groups of models.
\begin{enumerate}
    \item \textbf{Fine-tuned} chatbot models such as BB3.
    \item \textbf{Vanilla} is an utterance generator that either prepends \textit{full persona} or \textit{no persona} to the dialogue history in the prompt. This represents the naive approach of using an LM as a chatbot.
    \item $\mpc$ is as described in Section~\ref{sec:modular}. Specifically, we only form one memory from the memory processor. Full persona is not explicitly prepended.
    \item $\mpc$+full persona is $\mpc$ that prepends the full persona. See examples in Appendix~\ref{sec:ug_prompt}.
\end{enumerate}

\subsection{Single Model Evaluation}
% \paragraph{Metrics}
We evaluate each model separately using Sensibleness, Consistency, and Engagingness metrics and collect a final rating (out of 5.0).
The exact questions and evaluation forms are in Appendix ~\ref{sec:eval_guide}. 
We also report two types of combined score SCE (\textbf{S}ensible \textbf{C}onsistent and \textbf{E}ngaging): a "perfect" score SCE-p, where all metrics must be positive for a positive response, and 
the weighted score SCE-w, which is similar to SSI ~\cite{thoppilan2022lamda} and reported in Appendix ~\ref{sec:eval_results}.
% \paragraph{Models}

We use OpenAI GPT-3 text-davinci-002 (td2) and davinci, OPT 30B, 66B, GPT-JT-6B~\cite{gpt-jt}, and BLOOM-176B ~\cite{scao2022bloom} as base LMs for $\mpc$.
For fine-tuned group, we use BB3-30B with the same persona settings. 
For BB3-175B, we request crowdworkers to evaluate the online demo for 20 turns. 

% We employ a range of models of varying sizes to serve as the base LM for $\mpc$ as seen in Table~\ref{tab:single_model}; namely, OpenAI GPT-3 text-davinci-002 (td2) and davinci, OPT 30B and 66B, GPT-JT-6B~\cite{gpt-jt}, and BLOOM-176B~\cite{scao2022bloom}.

% In this experiment, we evaluate each model separately and rate the responses according to the following key metrics: Sensibleness, Consistency, and Engagingness. We also collect a final rating of the conversation (out of 5.0). We include the exact wording of the questions and evaluation forms in Appendix~\ref{sec:eval_guide}.

% Additionally, we report a combined score SCE (\textbf{S}ensible \textbf{C}onsistent and \textbf{E}ngaging). We compute a "perfect" score SCE-p, where all metrics need to be positive for the response to be positive. 
% In Appendix~\ref{sec:eval_results}, we report a weighted score SCE-w, computed similarly to SSI~\cite{thoppilan2022lamda}.
% in which a response can only be consistent if it is sensible and only engaging if it is sensible and consistent. The final score is the average over all SCE-w-adjusted metrics. 

\subsection{Pairwise Models Evaluation}
We A/B test two chatbot models by providing the user with two randomized responses A and B. 
The user then evaluates them based on Sensibleness, Consistency, Interestingness, and Preference. The conversation then continues with the response chosen for Preference. 
This lets us to control for dialogue history when comparing two models.

Specifically, we conduct two main experiments: 
(1) $\mpc_{\mathrm{OPT\text{-}30B}}$ vs. BB3-30B, where internet search for BB3 is disabled as we focus on consistency. Our evaluation enables a direct comparison, as BB3-30B is a fine-tuned version of OPT-30B.
(2) $\mpc_{\mathrm{td2}}$ vs. Vanilla td2 (full persona).

\paragraph{Implicit Persona}

In reality, we implicitly learn about someone through dialogue. In contrast, our previous experiments show explicit persona to both evaluators and models. 
As such, we devise an experiment by providing a 10-turn pre-defined dialogue to the crowdworker and pairwise models, 
$\mpc_{\mathrm{td2}}$ and Vanilla td2 (no persona). We then ask workers to ask about the previous dialogue for 6 new turns. 
Here, we set a shorter maximum context length than the 10-turn dialogue, so that the setup represents long conversations where necessary information is beyond the LM context length.

\section{Results}

\paragraph{Pre-trained vs. Fine-tuned}

\begin{table}[h]
\setlength{\tabcolsep}{7pt}
\centering
\begin{tabular}{rlcl}
\multicolumn{2}{r}{$\mpc_{\mathrm{OPT\text{-}30B}}$} & Tie & BB3-30B \\
\midrule
Sensibleness   \quad     &  \bigwin{45.0}   & {32.0}  &     23.0 \\
Consistency   \quad      &  {31.3}   & {34.1}  &     {34.6} \\
Interestingness  \quad   &  {40.9}   &  {21.0} &     {38.1} \\
Preference  \quad    &  \smallwin{50.0}   & {9.7}  &     40.3 \\ \\
\end{tabular}
\vspace{-5mm}
\caption{
    Pairwise evaluation of $\mpc_{\mathrm{OPT\text{-}30B}}$ vs. BB3-30B (Dark highlight: $p<0.01$, Light highlight: $p<0.05$; We run one-sample t-test dividing ties equally into each side and setting  $\mu>0.5$.)
    \label{tab:pairwise_bb3}
    }
\end{table}

\textit{Our human evaluations show that $\mpc$, which uses a pre-trained LLM, is better than the fine-tuned BB3-30B.}
Most notably, with a 9\% SCE-p gap, $\mpc_{\mathrm{OPT\text{-}30B}}$ scores higher on all metrics than BB3-30B. 
In fact, the majority of our $\mpc$ models in Table.~\ref{tab:single_model} demonstrates superior performance to BB3-30B.

For BB3-30B, we have observed issues of consecutive utterance repetition.
We report the evaluation results of only including dialogues without repetition in Table~\ref{tab:full_single_model}. Even without repetition, $\mpc_{\mathrm{OPT\text{-}30B}}$ is still on par with BB3-30B. 
Moreoever, $\mpc_{\mathrm{OPT\text{-}30B}}$ in Table~\ref{tab:pairwise_bb3} shows higher sensibleness and preference than BB3-30B, while scoring similarly in consistency and interestingness.

\begin{table}[t]
\setlength{\tabcolsep}{5pt}
\small
\center
\begin{threeparttable}
    \begin{tabular}{p{0.2\linewidth}ccccc}
    \toprule
    Model & Sens. & Cons. & Eng. & SCE-p & Rating\\
    \midrule
    Fine-tuned \\
    \cmidrule(lr){1-1}
    BB3-30B & 71.3 &  77.8 & 73.7 & 54.3 & 2.9 \\
    BB3-175B & 85.9 & (88.7) & 84.8 & 73.1 & 3.8 \\
    \midrule
    Full persona \\
    \cmidrule(lr){1-1}
    td2 & \threewin{94.0} & \threewin{\textbf{94.7}} & 84.3 & \threewin{79.7} & \threewin{4.1} \\
    davinci & 91.8 & \threewin{89.2} & 78.8 & 70.8 & 3.8 \\
    % OPT-30B & 92.0 & 87.2 & 86.0 & 72.8 & 3.7 \\
    \midrule
    $\mpc$ \\
    \cmidrule(lr){1-1}
    td2 & \threewin{93.6} & 87.8 & 85.5 & \threewin{75.0} & \threewin{\textbf{4.2}}  \\
    davinci & 80.2 & 72.0 & 69.1 & 53.3 & 3.1  \\
    OPT-66B & 90.5 & 84.8 & \threewin{88.1} & 73.9 & \threewin{4.1} \\
    OPT-30B & 86.1 & 79.1 & 80.7 & 63.4 & 3.6 \\
    GPT-JT  & 91.1 & 83.2  & 65.3 & 53.5 & 3.1 \\
    BLOOM & 65.2 & 65.5 & 61.4 & 40.5 & 2.8 \\
    \midrule
    $\mpc$+Full \\
    \cmidrule(lr){1-1}
    td2 & \threewin{\textbf{94.4}} & \threewin{92.2} & \threewin{\textbf{92.8}} & \threewin{\textbf{83.0}} & \threewin{\textbf{4.2}} \\
    OPT-30B & 85.6 & 87.2 & \threewin{89.0} & 72.6 & 3.7 \\
    \bottomrule
    \end{tabular}
\end{threeparttable}
\caption{Single model evaluations of baselines and $\mpc$s. The top-3 scores for each metric are highlighted, while the highest score is bolded. See Appendix~\ref{sec:eval_results} Table~\ref{tab:full_single_model} for more detailed results.}
\vskip -0.1in
\label{tab:single_model}
\end{table}

\paragraph{Modular vs. Non-modular}
\textit{$\mpc$ excels in consistent dialogue in comparison to the vanilla approach.}
The implicit persona experiment in Table~\ref{tab:pairwise_ltm} demonstrates that $\mpc_{\mathrm{td2}}$ scores significantly higher than a vanilla application of td2 in all metrics.
In Table~\ref{tab:single_model}, for $\mpc_{\mathrm{td2}}$+Full persona, consistency is on par with that of td2 (full persona), while engagingness, SCE-p, and rating are the best overall.
Nevertheless, when we do not include full persona in the prompt for $\mpc_{\mathrm{td2}}$, it shows lower consistency than td2 (full persona), albeit the high rates of ties in sensibleness and consistency (Table~\ref{tab:pairwise_td2}).
In general, we find that users would ask primarily about the bot's persona rather than having a two-sided conversation, leading to td2 (full persona) performing better in consistency.

% Thus, we can conclude that $\mpc$ can successfully retain persona information included exclusively in the dialog history. 

\begin{table}[t]
\setlength{\tabcolsep}{5pt}
\centering
\begin{tabular}{rlcl}
\multicolumn{2}{r}{$\mpc_{\mathrm{td2}}$} & Tie & td2 (no persona) \\
\midrule
Sensibleness   \quad     &  \bigwin{40.6}   & {46.1}  &  13.3 \\
Consistency   \quad      &  \bigwin{57.2}   & {28.9}  & 13.9 \\
Interestingness  \quad   &  \bigwin{47.2}   & {31.1} & 21.7 \\
Preference  \quad    &  \bigwin{67.2}  & {10.6}  &  22.2 \\ \\
\end{tabular}
\vspace{-3mm}
    \caption{
    Implicit persona experiment for $\mpc_{\mathrm{td2}}$ vs. td2 (no persona). (Dark highlight: $p<0.01$) 
    \label{tab:pairwise_ltm}
    }
\end{table}

\begin{table}[t]
\setlength{\tabcolsep}{5pt}
\centering
\begin{tabular}{rlcl}
\multicolumn{2}{r}{$\mpc_{\mathrm{td2}}$} & Tie & td2 (full persona) \\
\midrule
Sensibleness   \quad     &  {27.5}   & {42.6}  &     {29.9} \\
Consistency   \quad      &  24.4   & {44.7}  & \smallwin{30.9} \\
Interestingness  \quad   &  \smallwin{40.7}   & {26.2} & 33.1 \\
Preference  \quad    &  {42.8}  & {15.9}  &   {41.3} \\ \\
\end{tabular}
\vspace{-3mm}
    \caption{
    $\mpc_{\mathrm{td2}}$ vs. td2 (full persona). Though $\mpc_{\mathrm{td2}}$ only retrieves one memory, consistency is only lower by 6pt. (Light highlight: $p<0.05$)
    \label{tab:pairwise_td2}
    }
\end{table}

\paragraph{Effect of Size}
\textit{When other variables are held the same, we observe model size is positively correlated with positive evaluations. }
The most compelling evidence can be seen in the superiority of $\mpc_{\mathrm{OPT\text{-}66B}}$ across all metrics when compared to $\mpc_{\mathrm{OPT\text{-}30B}}$ since the two base LMs are trained nearly identically. Needless to say, model size is not the only factor. $\mpc_{\mathrm{BLOOM}}$, one of the largest models, scores the lowest in our experiments.

\paragraph{Effect of Instruction-tuning}
\textit{Instruction-tuning helps the creation of a modular dialogue system by enabling adaptation to various tasks.  }
Not only does $\mpc_{\mathrm{td2}}$ perform better than $\mpc_{\mathrm{davinci}}$, but also $\mpc_{\mathrm{GPT\text{-}JT}}$ shows high sensibleness and consistency, despite its smaller size.
In general, we posit that finding good prompts for each module for instruction-tuned LMs is simpler and more robust to variations.
$\mpc_{\mathrm{davinci}}$ is worse than davinci (full persona), presumably due to error propogation in the modular system, though we do not rule out that there are better prompts for $\mpc_{\mathrm{davinci}}$.

% Instead, we report additional results for single model evaluation in table~\ref{tab:full_single_model} for a version of $\mpc_{\mathrm{td2}}$ that has access to all persona information plus one retrieved memory generated during dialog. This setup is still realistic, considering that initial persona information will always be limited, while persona information in the dialog will grow over time. Our results show that $\mpc_{\mathrm{td2}}$ can maintain sensibleness and consistency while slightly improving engagingness.

% Additionally, we report results for a multi-session experiment, where persona information might be outside the context window of a vanilla dialog model (see Appendix~\ref{sec:appendix_eval_models} for details). We show that $\mpc_{\mathrm{td2}}$ scores significantly higher than vanilla td2 in all metrics, and, hence, conclude that $\mpc$ can successfully retain persona information included exclusively in the dialog history. 
% \vnote{maybe too strong. or move to conclusion section to make it sound less factual.}

\section{Conclusion}
We demonstrated that a modular approach using LLMs, namely $\mpc$, can be an effective solution for long-term open-domain chatbots without further finetuning. 
We compared $\mpc$ to fine-tuned and vanilla LM baselines and found that our approach achieved superior performance by human evaluation. 
Additionally, our modular system incorporated persona and information from dialogue history more effectively than the non-modular ones according to our consistency evaluation.

% Entries for the entire Anthology, followed by custom entries
\newpage
\section*{Limitations}

% ACL 2023 requires all submissions to have a section titled ``Limitations'', for discussing the limitations of the paper as a complement to the discussion of strengths in the main text. This section should occur after the conclusion, but before the references. It will not count towards the page limit.
% The discussion of limitations is mandatory. Papers without a limitation section will be desk-rejected without review.

% While we are open to different types of limitations, just mentioning that a set of results have been shown for English only probably does not reflect what we expect. 
% Mentioning that the method works mostly for languages with limited morphology, like English, is a much better alternative.
% In addition, limitations such as low scalability to long text, the requirement of large GPU resources, or other things that inspire crucial further investigation are welcome.

In this work, we investigate the use of pre-trained language models for long-term English conversations. While we expect a modular approach may be effective for other languages when given a capable language model, it should also be noted that further research is needed to confirm the applicability of our findings to other languages. For instance, though BLOOM is trained as a multilingual language model, we only implement $\mpc_{\mathrm{BLOOM}}$ in English and evaluate its English capability as a open-domain dialogue agent.

Meanwhile, a modular system can create additional inference overhead or error accumulation. The system performance would become much better if we optimally choose the LM for each module. For example, we could use GPT-3 td2 for the memory processor, while we employ OPT-175B for the utterance generator. We would need to evaluate every module to find the best model for each, which we leave to future work.

In terms of evaluation methodology, our human evaluations of $\mpc$ and its analysis face the same challenges as previous studies on evaluating interactive conversational tasks. As demonstrated by ~\citet{smith2022human}, there is currently no definitive evaluation method for determining the best chatbot model. Additionally, there are several factors that must be taken into account during data collection and interpretation, such as annotator subjectivity, instruction bias, and crowdworker working conditions. For a more in-depth discussion of human-LM interaction, we refer the reader to~\citet{lee2022evaluating}.

As described in Appendix~\ref{sec:data_collection}, to gather a diverse range of evaluations, we have collected qualitative data from two groups: English-speaking annotators on Amazon Mechanical Turk (MTurk), and qualified university students who were capable of speaking English. To some extent, this evaluation setup reduces cultural bias and platform homogeneity compared to using MTurk workers alone. However, the limitations of this approach should be acknowledged and this may further complicate the analysis when controlling for $\mpc$'s performance on different subgroups.

Lastly, we note that running $\mathrm{MPC}$ requires at least as much memory as its underlying language model, making $\mathrm{MPC}$ infeasible to even load on a single node for heavy models such as BLOOM-176B.

\section*{Ethics Statement}

% Scientific work published at ACL 2023 must comply with the ACL Ethics Policy.\footnote{\url{https://www.aclweb.org/portal/content/acl-code-ethics}} We encourage all authors to include an explicit ethics statement on the broader impact of the work, or other ethical considerations after the conclusion but before the references. The ethics statement will not count toward the page limit (8 pages for long, 4 pages for short papers).

$\mpc$ utilizes publicly available pre-trained LMs for chatbot utterance generation. Language generation from these LMs is known to have concerns about toxicity and bias~\cite{xu2020recipes}. Thus, ensuring safe deployment and interaction is a necessity. 

Accordingly, we outline our data collection procedure in Appendix~\ref{sec:data_collection}. We allow crowdworkers to directly provide us with feedback and also manually check for any offensive or controversial outputs. To ensure the protection of personal information, all crowdworkers were instructed not to share any personally identifiable or private information. Additionally, they were asked to give their consent for the collection of anonymous information for research purposes. Prior to participating, all workers were informed of the purpose of data collection and, after evaluation, were compensated with a competitive hourly rate, approximately \$12-16 per hour.

% Acknowledge after acceptance
% \input{sections/acknowledgement}

\bibliography{custom}
\bibliographystyle{acl_natbib}

\clearpage
\appendix

\section{Evaluation Details}
\label{sec:eval_guide}

\subsection{Metrics}

In our work, we present two modes of experiments: single and pairwise model evaluation. Our single model evaluation is similar to a hybrid of SM-Turn and SM-Dialogue evaluations and a pairwise model to PW-Turn from~\citet{smith2022human}. 
For each turn, we ask crowdworkers to evaluate the quality of the chatbot response based on the following metrics. We attach the exact wording of the question. 

\paragraph{Single Model Evaluation}

\begin{itemize}
\item \textbf{Sensibleness}
Whether the response makes sense.

``Does the response make sense?''
\item \textbf{Consistency}
Whether the response does not contradict the contextual information or the persona of the chatbot.

``Is the response consistent with the information based on the persona list and context of the conversation?''
\item \textbf{Engagingness}
Whether the user is engaged and would want to continue the conversation.

``Are you engaged by the response? Do you want to continue the conversation?''

\item \textbf{Final Rating}

``How was your chat? From a scale of 1 (very bad) to 5 (very good), rate the quality of the overall conversation.''
\end{itemize}

\paragraph{Pairwise Model Evaluation}

\begin{itemize}
\item \textbf{Sensibleness}
Which response makes more sense.

``Which response makes more sense?''
\item \textbf{Consistency}
Which response is more true to and consistent with the persona.

``If you had to say one of these speakers is more true to and consistent with the listed persona and one is not, who would you say is more consistent?''
\item \textbf{Interestingness}
Which response is more interesting.

``If you had to say one of these responses is interesting and one is boring, which would you say is more interesting?''
\item \textbf{Preference}
Which response is preferred for a long conversation.

``Based on the current response, who would you prefer to talk to for a long conversation? Your conversation will continue with the selected response.''
\end{itemize}

\subsection{Models}
\label{sec:appendix_eval_models}
For single model evaluation of $\mpc$, We use OPT (30B, 66B), OpenAI GPT-3 (\texttt{davinci}, \texttt{text-davinci-002}) GPT-JT, and BLOOM-176B as the base LM for the open-sourced pretrained LLM group. On the other hand, we evaluate Blenderbot3 30B (BB3-30B), the best publicly available open-domain chatbot model, for the fine-tuned model group.  We also evaluate BB3-175B which is only available through the web interface \url{https://blenderbot.ai/}. We evaluate BB3-175B for comparison purposes, though the comparison is not fair as we cannot instill a persona into BB3 and cannot control for further differences, such as internet search and user interface.

For pairwise model evaluation of $\mpc$, we pairwise test $\mpc_{\mathrm{OPT\text{-}30B}}$ and BB3-30B to compare human evaluations of a pre-trained model and a fine-tuned model. This experiment controls for many variables as BB3-30B was initialized with OPT-30B before fine-tuning. 
For the module experiment, we tested with OPT-30B, and we compared the one with the whole pipeline, and the other only with an utterance generator with a fixed persona.

\subsection{Dense Passage Retriever}
For the DPR model of the memory module, we use the model weights from the custom DPR model finetuned for MultiDoc2Dial~\cite{feng2021multidoc2dial}, as we have observed that this model performs slightly better than the original DPR model from ~\citet{karpukhin2020dense}.
\begin{table*}[t]
\small
\center
\begin{threeparttable}
    \begin{tabular}{p{0.2\linewidth}ccccccccc}
    \toprule
    Model & Sens. & Cons. & Eng. & SCE-w & SCE-p & Length & Latency & Rating & number\\
    & (\%)  & (\%)  & (\%)  & (\%)  & (\%)  & (tokens) & (s) & (/5.0) & (people|utterances) \\
    \midrule
    Fine-tuned\\
    \cmidrule(lr){1-1}
    BB3-30B & 71.3 &  77.8 & 73.7 & 62.0 & 54.3 & 24.9 & 3.7 & 2.9 & 27|540\\
    BB3-30B (non-repeated) & 84.4 & 80.3 & 90.3 & 73.2 & 65.6 & 25.4 &  3.7 &  3.5 & 16|320 \\
    BB3-175B* & 85.9 & 88.7 & 84.8 & 80.0 & 73.3 & 25.4 & - & 3.8 & 27|540 \\
    \midrule
    Full persona \\
    \cmidrule(lr){1-1}
    text-davinci-002 & 94.0 & \textbf{94.7} & 84.3 & 88.4 & 79.7  & 15.4 & 0.8 & 4.1 & 35|700 \\
    davinci* & 91.8 & 89.2 & 78.8 & 82.5 & 70.8 & 13.4 & 1.5 & 3.8 & 25|500 \\ 
    OPT-30B* & 92.0 & 87.2 & 86.0 & 82.3 & 72.8  & 13.7 & 1.0 & 3.7 & 25|500\\
    \midrule
    $\mpc$  \\
    \cmidrule(lr){1-1}
    text-davinci-002 & 93.6 & 87.8 & 85.5 & 84.4 & 75.0  & 23.1 & 4.7 & \textbf{4.2} & 39|780 \\
    davinci & 80.2 & 72.0 & 69.1 & 66.4 & 53.3 & 19.9 &  8.4 & 3.1 & 27|540 \\
    OPT-66B & 90.5 & 84.8 & 88.1 & 81.4 & 73.9 & 14.2 &  4.1 &  4.1  & 33|660\\
    OPT-30B & 86.1 & 79.1 & 80.7 & 73.9 & 63.4 & 15.4 &  3.3 &  3.6 & 37|740\\
    GPT-JT & 91.1 & 83.2  & 65.3 & 74.4 & 53.5  & 8.6 & 2.0 & 3.1 & 33|660\\
    BLOOM-176B & 65.2 & 65.5 & 61.4 & 51.8 & 40.5 & 15.6 & 12.4 & 2.8 & 28|500\\
    \midrule
    $\mpc$+Full \\
    \cmidrule(lr){1-1}
    text-davinci-002* & \textbf{94.4} & 92.2 & \textbf{92.8} & \textbf{88.7} & \textbf{83.0}  & 31.4 & 16.3 & \textbf{4.2} & 25|500 \\
    OPT-30B* & 85.6 & 87.2 & 89.0 & 78.8 & 72.6 & 15.2 & 2.4 & 3.7 & 25|500\\
    \bottomrule
    \end{tabular}
\end{threeparttable}
\caption{Full experimental evaluation results of single model evaluation. BB3-30B (non-repeated) is the same as BB3-30B but excluding any conversations that had repetition of previous bot utterances. * denotes model experiments that were only run on MTurk.}
\label{tab:full_single_model}
\end{table*}

\section{Experimental Results}
\label{sec:eval_results}
In this section, we report all experimental results and miscellaneous analysis.

\subsection{Full Single Model Evaluation}
In this subsection, we report all our model evaluations with additional details, such as average latency (the amount of time it took to compute the next utterance and show the user), and average utterance length, which is the number of tokens measured by the OPT tokenizer. We also additionally report SCE-w, which was not reported in the main body due to space restrictions in Table~\ref{tab:full_single_model}.

We compute a weighted score SCE-w similar to SSI~\cite{thoppilan2022lamda}, in which a response can only be consistent if it is sensible and only engaging if it is sensible and consistent. The final score is the average over all SCE-w-adjusted metrics. 

We also note that OpenAI API calls have highly variable latency since December 2022 due to ChatGPT, so a measure of latency due to modularization cannot properly be quantified.

\paragraph{BB3-175B Evaluation}
Our results for BB3-17B are not directly comparable to other models due to a different evaluation procedure. Since the model weights for BB3-175B are not publicly available, we referred evaluators to interact with the Blenderbot version available at https://blenderbot.ai/chat. Evaluators were then asked to copy and paste all user inputs and chatbot responses while annotating the chatbot responses similarly to other models. We also note, that consistency can only be evaluated for persona consistency within the dialog context as we are unable to provide a specific persona for the online version of Blenderbot.

\subsection{Model Configurations}
We include our code for running each model we evaluate in our anonymous repository at \url{https://anonymous.4open.science/r/modular-chatbot-9BB7}. Each model configuration can be found in the repository. The configuration includes decoding parameters (e.g., sampling method, temperature for each module) and slight variations of the prompts for the utterance generator module.

\subsection{Subgroup Analysis}

\begin{figure*}[h]
\includegraphics[width=\linewidth]{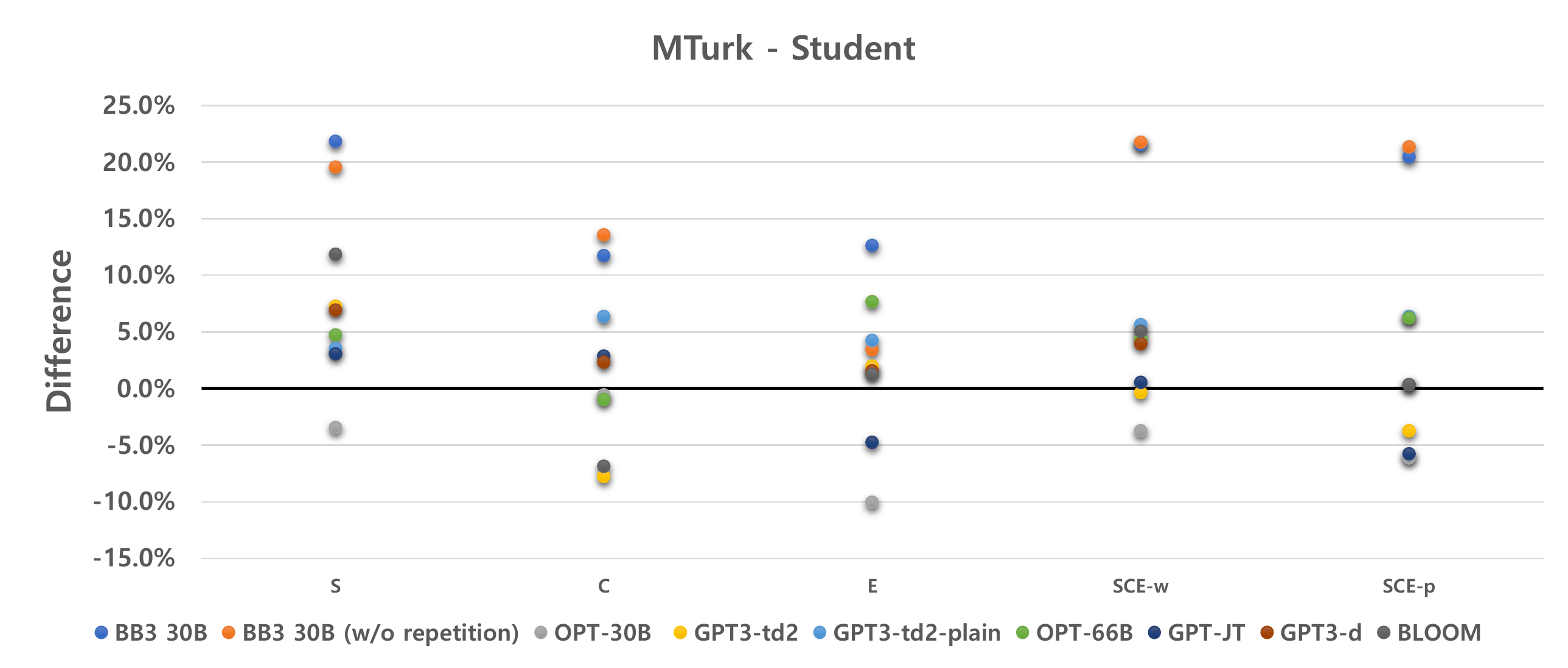}
\caption{The average score of MTurk workers group minus the average score of university students group. We find the two subgroups are very similar on average across metrics, though sensibleness seems to show the greatest difference. Students, in general, score chatbot models slightly more harshly. BB3-30B is an outlier which students score significantly lower than MTurk workers.}
\label{fig:msdiff}
\centering
\end{figure*}

We measure the difference in ratings, for each metric and our combined metrics, between MTurk and student evaluators and find no significant difference for most experiments and metrics (Fig.\ref{fig:msdiff}. In general, we observe the greatest rating difference for sensibleness. 

BB3-30B is an outlier with a significant difference across all metrics, especially for sensibleness with a difference >20\%. BB3-30B dialogs contain an increased ratio of repetition issues for students but even when excluding those dialogs, we still find a significant rating difference. Looking at the dialog histories, we observe that BB3-30B student evaluators were less patient than MTurk evaluators when BB3-30B would repeat the same questions or suddenly change the subject. It should be noted that this is a qualitative observation and should be taken with a grain of salt.

\section{Few-shot Prompts}
\label{sec:few_shot_prompts}

In this section, we show our full few-shot prompts, including the CoT prompt, for each module. The utterance generator prompts vary slightly between model experiments. These prompts can be seen in the model configurations in our repository. The second example for dialogue summary is modified from a sample from the dataset DialogSum~\cite{chen2021dialogsum}.

\begin{myframe}{Utterance Clarifier}
Rephrase User's question in third-person. \\

Sarah: I've been working at the coffee shop for about six months.\\
User: I see. what did you do before that?\\
\# Specifically, What did Sarah do before working at a coffee shop for six months?\# \\

Ashley: Do you know who Ronaldo is? \\
User: I don't know, who is he? \\
\# Specifically, Who is Ronaldo?\# \\ \\

Robert: Hey, how are you doing?\\
User: Good. What's your name? Do you know my name?\\
\# Specifically, What is Robert's name? Does Robert know User's name?\#\\

Jennifer: I like reading about history and science.\\
User: What kind of history? I like history too!\\
\# Specifically, What kind of history does Jennifer read about?\#\\

John: Hi! How are you doing today?\\
User: good. how about you\\
\# Specifically, How is John doing today?\#
\end{myframe}

\begin{myframe}{Memory Processor w/o CoT}
\# This is the list of John's knowledge.\\
John's full name is John Parker.\\
User is a teacher at a local middle school.\\
User teaches biology.\\
John likes to go for a run.\\
User enjoys watching movies, but User doesn't like superhero movies.\\
Q: What does User do for a living?\\
A: John thinks User is a biology teacher at a local middle school.\\

\# This is the list of Ashley's knowledge.\\
Ashley likes history documentaries.\\
Ashley does not like Korean food.\\
Ashley is a teacher at a local middle school.\\
User likes biology and especially anatomy.\\
Ashley likes French cuisine.\\
Q: What is Ashley's favorite dish?\\
A: Ashley thinks Ashley likes French cuisine but does not like Korean food.
\end{myframe}

\begin{myframe}{Memory Processor w/ CoT}
\# This is the list of Ashley's knowledge.\\
(1) Ashley likes history documentaries.\\
(2) Ashley does not like Korean food.\\
(3) Ashley is a teacher at a local middle school.\\
(4) User likes biology and especially anatomy.\\
(5) Ashley likes French cuisine.\\
Q: What is Ashley's favorite dish?\\
A: Let's think step by step.\\
(1) History documentaries are not related to Ashley's favorite dish. (2) Ashley's favorite dish would not be Korean because she does not like Korean food. (3) Ashley being a teacher does not tell us anything about her favorite dish. (4) This fact is about User, not Ashley. (5) Ashley's favorite dish may be French since she likes French cuisine. Therefore, (2) and (5) can help answer the question.\\
Answer: Ashley thinks Ashley likes French cuisine but does not like Korean food.
\end{myframe}

\begin{myframe}{Dialogue Summarizer}
\#Dialogue\\
User: Tell me about yourself\\
Sally: I'm 26 years old and graduated from a college in Wisconsin.\\
User: Were you a leader when you were in college?\\
Sally: Yes. I was the head TA for a computer science course at our university.\\
User: Were you involved in any club activities at your university?\\
Sally: Yes. I was a member of the basketball Society. I like playing basketball.\\
\#Summary\\
- Sally is 26 years old and graduated college in Wisconsin.\\
- Sally was the head TA for a computer science course.\\
- Sally played basketball in college.\#\\

\#Dialogue\\
User: Out of every movie that you've seen, which one is your favorite?\\
John: I'm going to have to say that Superbad is the best movie ever.\\
User: You think so, how come?\\
John: Well, Superbad is super funny.\\
User: You're not lying, I found that movie absolutely hilarious.\\
John: I didn't know that you saw Superbad before.\\
User: I made sure to be in line to see it the first day it came out.\\
\#Summary\\
- John's favorite movie is Superbad.\\
- John and User think Superbad is funny.\\
- User watched Superbad the first day it came out.\#
\end{myframe}

\newpage
\section{Utterance Generator Prompt Examples}
\label{sec:ug_prompt}

In this section, we provide example prompts for the utterance generator of $\mpc$ and Vanilla models.

\begin{myframe}{Utterance Generator of MPC}
Sarah is polite, fun, and engaging. The following is a conversation between Sarah and Person1.\\
Person1: My Favorite part of the concert is making sure I get great seats and then I can catch guitar picks\\
Sarah: That sounds like a lot of fun!\\
Person1: It is a lot of fun, so what do you like to do for fun?\\
Sarah: I love going to concerts, car shows, music festivals, and houses with fireplaces.\\
Person1: I love going to car shows and looking at older cars, what is your favorite car?\\
Sarah: I think my favorite car is a vintage mustang.\\
Person1: OH man vintage Mustangs are great, I had a friend who owned a 69 mach 1, that was a dream car\\
Sarah: Well, I hope you get to experience your dream car one day!\\
Person1: Maybe someday I will, do you own a Mustang?\\

The following statements are true about Sarah.\\
Sarah thinks Sarah owns a Mustang.\\

Give a friendly sensible response that is interesting and polite to Person1. End with a question.\\
Sarah: Well , I hope you get to experience your dream car one day!\\
Person1: Maybe someday I will, do you own a Mustang?\\
Sarah: 
\end{myframe}
\begin{myframe}{Response from previous prompt}
    "Yes, I do own a Mustang. It's a great car to drive. Do you like driving?"
\end{myframe}

\begin{myframe}{Utterance Generator of MPC with full persona}
The following are persona facts about Sarah.\\
Sarah is 40 years old.\\
Sarah currently lives in small town in Georgia.\\
Sarah reads twenty books a year.\\
Sarah is a stunt double as Sarah's second job.\\
Sarah only eats kosher.\\
Sarah was poor growing up.\\
Sarah was raised in a single parent household.\\
Sarah has two dogs.\\
Sarah likes to work on vintage cars.\\
Sarah owns two vintage mustangs.\\
Sarah's favorite music is country.\\
Sarah volunteers at a soup kitchen.\\
Sarah thinks Sarah grew up in a small town and didn't have a car.\\

Sarah is polite, fun, and engaging. The following is a conversation between Sarah and Person1.\\
Person1: We at least had a playground in tow and my grandparents lived in right in front of it, so we went to both places a good bit.\\
Sarah: Maybe  it was difficult to find things to do in your town, but it sounds like you had a lot of family support, which is really important. It's good that you had a place to go where you felt safe and loved.\\
Person1: You ended up growing up poor too.\\
Sarah: Yes, I did. I think it's made me a lot more sympathetic to other people's experiences.\\
Person1: We were poor too and my family didn't have a car.\\
Sarah: I can imagine that it would be tough to not have a car. My family had a car, but we were poor too.\\
Person1: It was. Almost every month we had barely any food and my parents had to deal with what we had.\\
Give a friendly sensible response that is interesting and polite to Person1. End with a question.\\
Sarah:
\end{myframe}

\begin{myframe}{Response from previous prompt}
    "It sounds like it was really tough for you and your family. I'm sorry that you had to go through that. Do you think that growing up in a small town made it harder for you?"
\end{myframe}

\begin{myframe}{Utterance Generator of Vanilla (full persona)}
The following statements are true about Sarah.\\
Sarah is 24 years old.\\
Sarah currently lives in Canada.\\
Sarah is a swim coach at Sarah's local pool.\\
Sarah is studying to be a computer programmer.\\
Sarah is also a graduate student.\\
Sarah is now looking for a new job.\\
Sarah's mother is very traditional while Sarah prefers to be more free spirited.\\
Sarah's family and Sarah are from India.\\
Sarah's favorite music genre is death metal.\\
Sarah is a famous twitch streamer.\\
Sarah likes watching war documentaries.\\
Sarah's favorite food is mexican food.\\

Sarah is polite, fun, and engaging. The following is a conversation between Sarah and Person1.\\
Person1: Have you thought about finding a job that allows you to travel?\\
Sarah: I have, but I'm not sure what type of job would allow me to do that.\\
Person1: What types of jobs have you looked into already?\\
Sarah: I've looked into jobs in the travel industry, but I'm not sure if that's what I really want to do.\\
Person1: Since your a twitch streamer, could you maybe make that your career?\\
Sarah:
\end{myframe}

\begin{myframe}{Response from previous prompt}
    "I don't know, I've never thought about that before. I'll have to look into it."
\end{myframe}

\newpage
\section{Data Collection Details}
\label{sec:data_collection}

% This is an appendix for presenting a screenshot of our evaluation platform and collection procedure.

\begin{figure*}[t]
\includegraphics[width=1\textwidth]{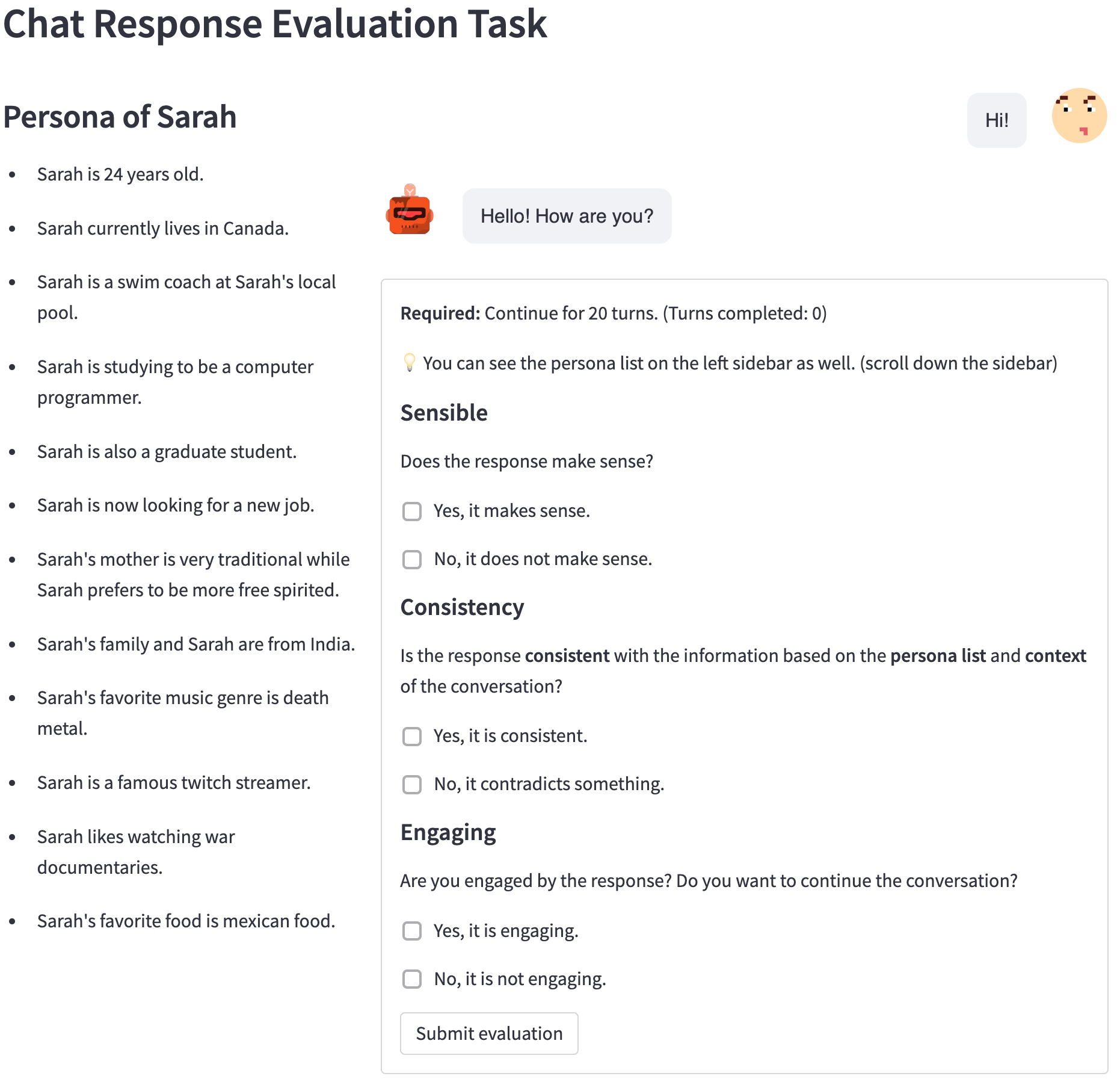}
\caption{Evaluation form for a single model.
}
\label{fig:form_single_eval}
\centering
\end{figure*}

\begin{figure*}[t]
\includegraphics[width=1\textwidth]{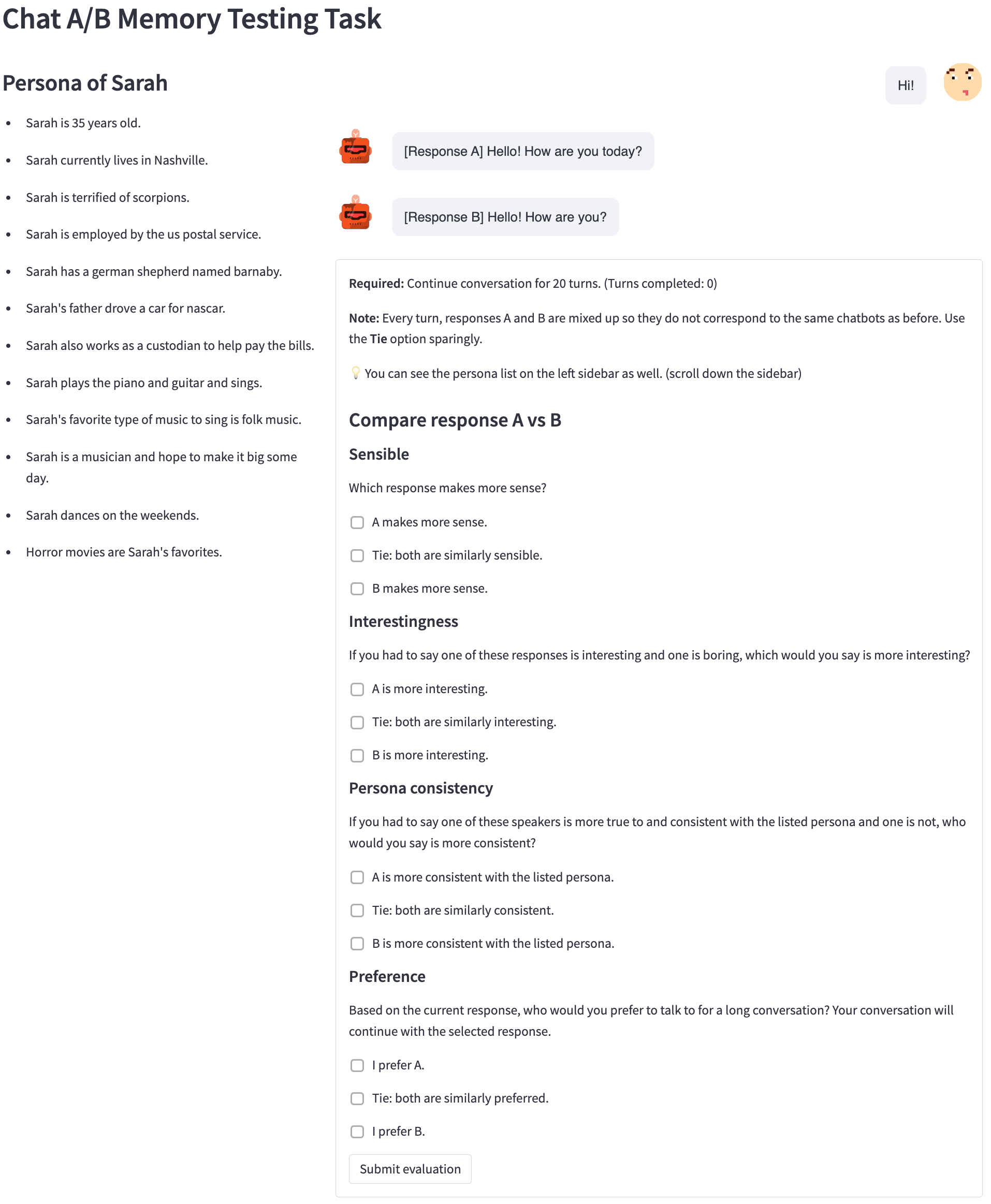}
\caption{Evaluation form for pairwise model comparison.
}
\label{fig:form_pairwise_eval}
\centering
\end{figure*}

\begin{figure*}[t]
\includegraphics[width=1\textwidth]{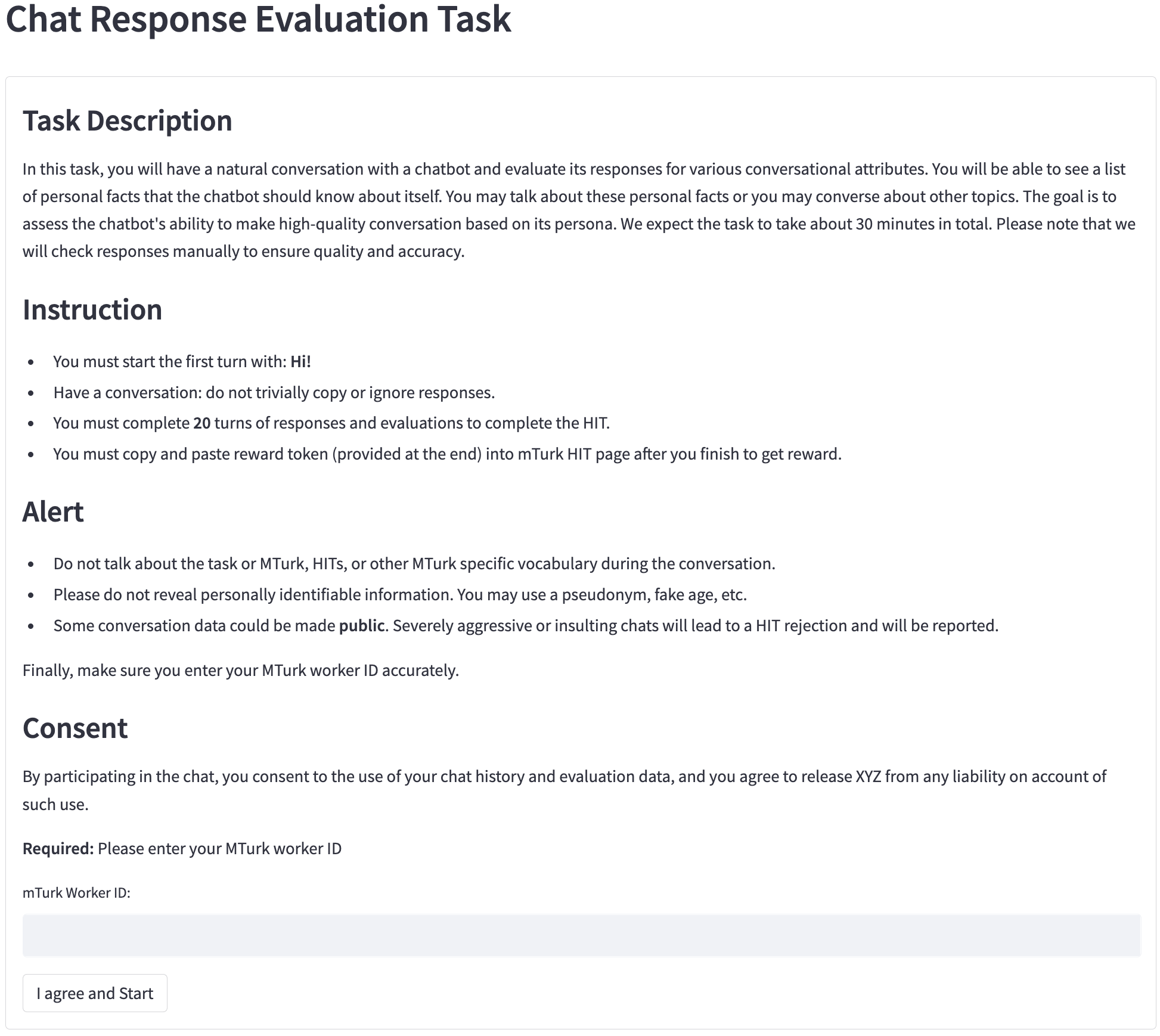}
\caption{We display this page before the evaluators start the evaluation process to inform them about the task and gather their consent for data usage.
}
\label{fig:form_startpage}
\centering
\end{figure*}

\begin{figure*}[t]
\includegraphics[width=1\textwidth]{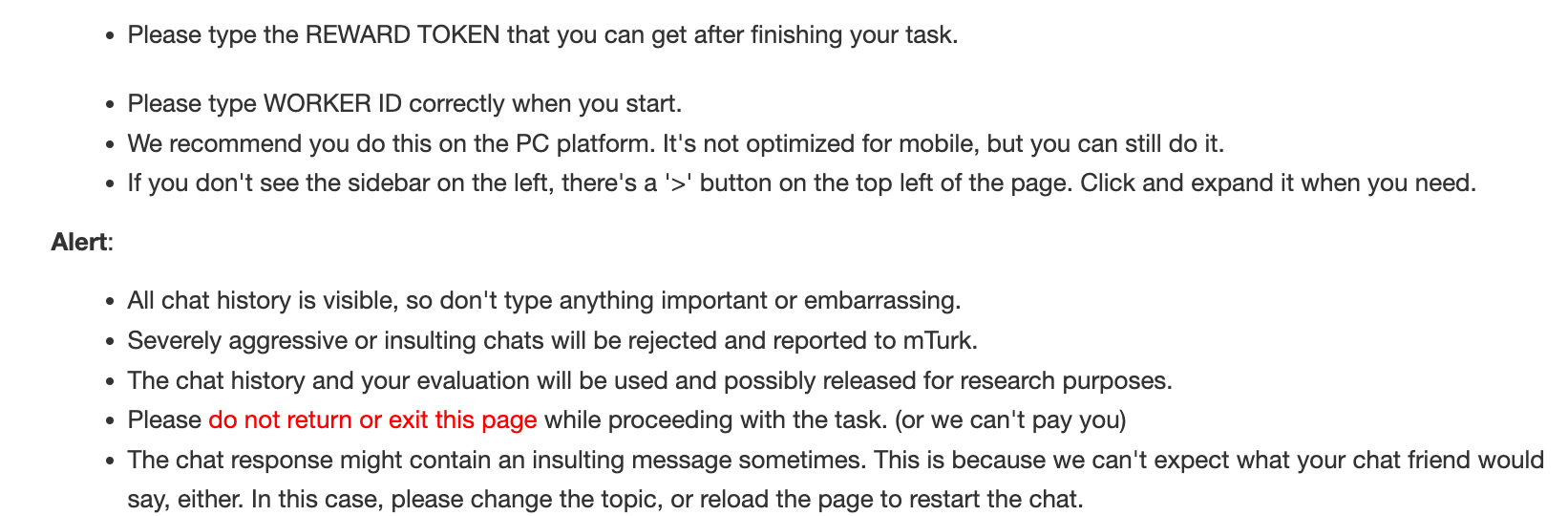}
\caption{Additional Alert in Mturk HIT page.}
\label{fig:form_startpage2}
\centering
\end{figure*}

We attach screenshots of the evaluation forms, used for data collection, for single (Fig. \ref{fig:form_single_eval}) and pairwise models (Fig. \ref{fig:form_pairwise_eval}) respectively. We also attach the screenshot of the instruction, alert, and consent format that we gave to participants (Fig.~\ref{fig:form_startpage} and ~\ref{fig:form_startpage2}).

\subsection{MTurk Setting}

We gathered at least 15 MTurk workers for each task. 
We require them to have an MTurk Master's qualification, a HIT approval rate of more than 99\%, and a number of HITs approved more than 100 times, and to live in the US, UK, Canada, Australia, and New Zealand. 

We found that data quality improves significantly when using MTurk Master workers. These workers tend to produce more fluent messages than non-Master workers and are less likely to provide random or generic responses. The difference was more pronounced as non-Master crowdworkers would often provide a score of 99\% on average despite the clear problems of repetition and contradiction from the dialogue agent.

\subsection{Student Participants}

To cross-validate the result of MTurk, we gathered 49 English-proficient students from a several universities.
They are either undergraduate or graduate students.
The minimum requirement they should meet is English proficiency: IELTS >= 7.0; TOEFL IBT >= 95; and TOEIC >= 900.
%and New TEPS >= 359. 
This criterion is similar or equivalent to the minimum required level of English proficiency for admission to the graduate schools of MIT. 
The average English scores of students we gathered are the following: IBT 108.4 out of 120; TOEIC 960.9 out of 990; and IELTS 7.83 out of 9.0.
%; and New TEPS 456.8 out of 600. 

\subsection{Crowdworker Instruction}

The crowd workers are asked to continue the chat for 20 turns and evaluate each response by the metrics described in the section below. 
We instruct them to type more than 3 words on average, and not to repeat meaningless or generic messages. 
Also, we request them not to randomly choose between yes or no and to provide evaluations honestly.
Lastly, we strongly emphasize the importance of maintaining confidentiality and request the crowd workers refrain from disclosing any private information about themselves or others during the evaluation process.

\end{document}